# Bipedal Walking Robot using Deep Deterministic Policy Gradient

Arun Kumar
*Aerospace Engineering Department*
*Indian Institute of Science*
Bangalore, India
arun.etc.kumar@gmail.com

Navneet Paul
*Aerospace Engineering Department*
*Indian Institute of Science*
Bangalore, India
nav74neet.paul@gmail.com

S N Omkar
*Aerospace Engineering Department*
*Indian Institute of Science*
Bangalore, India
omkar@iisc.ac.in

*Abstract*—Machine learning algorithms have found several applications in the field of robotics and control systems.The control systems community has started to show interest towards several machine learning algorithms from the sub-domains such as supervised learning, imitation learning and reinforcement learning to achieve autonomous control and intelligent decision making. Amongst many complex control problems, stable bipedal walking has been the most challenging problem. In this paper, we present an architecture to design and simulate a planar bipedal walking robot(BWR) using a realistic robotics simulator, Gazebo. The robot demonstrates successful walking behaviour by learning through several of its trial and errors, without any prior knowledge of itself or the world dynamics. The autonomous walking of the BWR is achieved using reinforcement learning algorithm called Deep Deterministic Policy Gradient(DDPG). DDPG is one of the algorithms for learning controls in continuous action spaces. After training the model in simulation, it was observed that, with a proper shaped reward function, the robot achieved faster walking or even rendered a running gait with an average speed of 0.83 m/s. The gait pattern of the bipedal walker was compared with the actual human walking pattern. The results show that the bipedal walking pattern had similar characteristics to that of a human walking pattern. The video presenting our experiment is available at https://goo.gl/NHXKqR.

*Index Terms*—Bipedal walking robot, Reinforcement learning, Deep Deterministic Policy Gradient, Simulation, Gazebo

## I. INTRODUCTION

Over the past three decades the robotics research community around the world has shown considerable interest in the field of humanoid robotics [1]–[3]. One of the main reasons for this interest is that we humans tend to interact or relate more with human-like entities [3], [4]. Also, the domain of legged robots for traversing uneven, unstable terrains have intrigued several roboticists. Bipedal walking robots are one typical classification of the humanoid robots that has garnered numerous research efforts in the past couple of decades. The legged based locomotion of the humanoid robots have a superior advantage over its conventional wheel-based counterparts, as it provides the possibility of either replacing or assisting humans in adverse environments [5]. Moreover robots that are biologically inspired or those modelled anthropomorphically render greater adaptability in diverse environments, especially, ones requiring human intervention and needs [2]. Ease of overcoming random obstacles while travelling in complex dynamic environments has been advantageous for the bipedal robots compared to other legged robots like quadrupeds, etc [6]. From a biomechanics research point of view, understanding biped stability and walking mechanism lays an important foundation for better understanding of how humans traverse from one place to another [7]. The human locomotion, although simple as it appears to be, is a highly complex manoeuvre involving multiple degrees of freedom that are in turn coupled with complex non-linear dynamics generated due to various extensor and flexor muscle groups in the lower body. This served as one of the main motivations for proper understanding of the physiology involved in human locomotion research and replicate the same on a BWR [7]. While the bipedal walking robots are known for their ease and flexibility of traversing over a wide range of terrains, the stability is the main concern.BWRs pose exceptional challenges and concerns to control systems and designs mainly due to their non-linearity and instability for which well developed classical control architecture cannot be applied directly. Discrete phase change from statistically stable double-stance position to the statistically unstable single-stance position in the dynamics of the BWR demand suitable control strategies [8]. Solving the stability issue of a bipedal walking system has aroused curiosity among many control scientists over the years [9], [10]. These conventional control theory approaches rely on the complex deterministic and mathematical engineering models. Zero Moment Point (ZMP) is one of the conventional method which is adopted as an indicator for dynamic stability in BWRs [11]. However, there are certain drawbacks associated with ZMP based control methods that involve energy inefficient walking, limited walking speed and poor resistance to external perturbations [12]. This method often relies on both high level of mathematical computations and perfect knowledge of both the robot and environment parameters [13], [14]. Several machine learning practises have emerged over the recent years that prove to have an edge over the conventional classical systems and control theory approaches to achieve stable bipedal walking. Reinforcement learning is a sub-domain of machine learning, that could be applied as a model-free learning of complex control systems [15]. Specifically, model-free learning of bipedal walking has mostly revolved around implementing several action policy learning algorithms based on Markov Decision Process (MDP) [16], [17]. Several state of the art reinforcement learning algorithms with the MDP have produced significant results when implemented in fully observable simulated environments [18]. This has motivated an increasing number of computer scientists and robotics researchers to exploit reinforcement learning(RL) methods to allow the agents to perform dynamic motor tasks in more complex and adverse environments [19], [20]. Our contributions in this research:

- Suggest a framework for implementing reinforcement learning algorithms in Gazebo simulator environment.
- Implement Deep Deterministic Policy Gradient based RL algorithm for efficient and stable bipedal walking.
- Compare the bipedal walker's gait pattern with an actual human's gait pattern.

The paper is organized as follows. In section II, a basic introduction to reinforcement learning, followed by the description of the DDPG algorithm is presented. Section III, describes the mechanical design of planar BWR model and the virtual environment simulator, Gazebo. Section IV elucidates the results of the simulation along with the discussions. Section V consists of the conclusions and future scope of research.

## II. REINFORCEMENT LEARNING

Humans rely on learning from interaction, repeated trial and errors with small variations and finding out what works and what not. Let's consider an example of a child learning to walk. It tries out various possible movements. It may take several days before it stands stably on its feet let alone walking. In the process of learning to walk, the child is punished by falling and rewarded for moving forward [8]. This rewarding system is inherently built into human beings that motivate us to do actions that garner positive rewards (e.g., happiness) and discourage the actions that account for bad rewards(e.g., falling, getting hurt, pain, etc.).

This kind of reward based learning has been modelled mathematically and is called Reinforcement learning. A typical reinforcement learning setting consists of an agent interacting with the environment $E$. After each time step $t$, the agent receives an observation $s_t$ from the environment and takes an action $a_t$. The agent receives a scalar reward $r_t$ and the next state $s_{t+1}$ [21][Fig.1].

The technique by which the agent chooses actions is called a policy of the agent, $\pi$ which maps the states to the probability distribution over the actions $\pi : S \to P(A)$. The environment can be modelled as an MDP with a state space $S$, action space $\mathcal{A} = \mathbb{R}^N$, an initial state distribution $p(s_1)$, transition dynamics $p(s_{t+1}|s_t,a_t)$, and reward function $r(s_1,a_t)$. The sum of discounted future reward is called the return from a state and is defined as $R_t = \sum_{i=t}^{T}\gamma^{i-t} r(s_i,a_i)$ with a discounting factor $\gamma \epsilon [0, 1]$. The goal of the agent is to maximize the expected return from the start distribution $J = \mathbb{E}_{r_i,s_i \sim E, a_i \sim \pi}[R_1]$. The action-value function describes the expected return after taking an action $a_t$ in state $s_t$ and following policy $\pi$ thereafter:

$$Q^\pi(s_t, a_t) = \mathbb{E}_{r_{i \geq t}, s_{i > t} \sim E, a_{i > t} \sim \pi}[R_t | s_t, a_t] \quad (1)$$

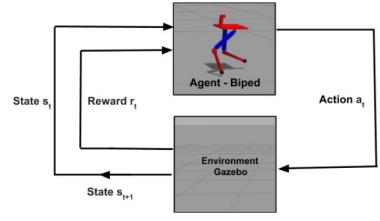

Fig. 1. Agent - Environment interaction in Reinforcement Learning

For a deterministic policy $\mu : S \to A$ we can write the Bellman Equation as:

$$Q^\mu(s_t, a_t) = \mathbb{E}_{r_t, s_{t+1} \sim E}[r(s_t, a_t) + \gamma Q^\mu(s_{t+1}, \mu(s_{t+1}))] \quad (2)$$

The off-policy algorithms like Q-Learning use the greedy policy $\mu(s) = \text{argmax}_a Q(s,a)$. Function approximators parameterized by $\theta^Q$, which is optimized by reducing the loss function:

$$L(\theta^Q) = \mathbb{E}_{s_t \sim \rho^\beta, a_t \sim \beta, r_t \sim E}[(Q(s_t, a_t|\theta^Q) - y_t)^2] \quad (3)$$

where

$$y_t = r(s_t, a_t) + \gamma Q(s_{t+1}, \mu(s_{t+1})|\theta^Q)$$

### A. Deep Deterministic Policy Gradient (DDPG)

DDPG is an actor-critic algorithm based on Deterministic Policy Gradient [21], [22]. The DPG algorithm consists of a parameterized actor function $\mu(s|\theta^\mu)$ which specifies the policy at the current time by deterministically mapping states to a specific action. The critic Q(s, a) is learned using the Bellman equation the same way as in Q-learning. The actor is updated by applying the chain rule to the expected return from the start distribution $J$ with respect to the actor parameters:

$$\nabla_{\theta^\mu} J \approx \mathbb{E}_{s_t \sim \rho^\beta}[\nabla_a Q(s, a|\theta^Q)|_{s=s_t, a=\mu(s_t|\theta^\mu)}]$$
$$= \mathbb{E}_{s_t \sim \rho^\beta}[\nabla_{\theta^\mu} Q(s, a|\theta^Q)|_{s=s_t, a=\mu(s_t)} \nabla_{\theta_\mu}\mu(s|\theta^\mu)|_{s=s_t}] \quad (4)$$

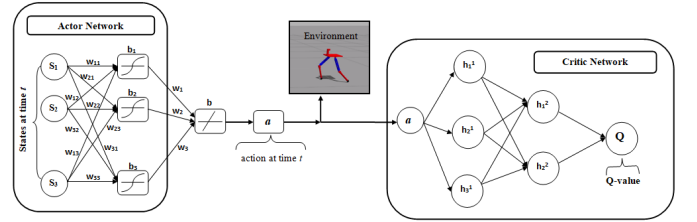

Fig. 2. Actor-Critic Network Model

DDPG combines the merits from its predecessor algorithms to make it more robust and efficient in learning. The samples obtained from exploring sequentially in an environment are not independently and identically distributed so, DDPG uses the idea from Deep Q-Networks(DQN) called replay buffer.

**Algorithm 1** Deep Deterministic Policy Gradient (DDPG)

1: Randomly initialize critic network Q(s,a|$\theta^Q$) and actor $\mu$(s|$\theta^Q$) with weights $\theta^Q$ and $\theta^\mu$.
2: Initialize the target network Q' and $\mu$' with weights
$$\theta^{Q'} \leftarrow \theta^Q, \theta^{\mu'} \leftarrow \theta^\mu$$
3: Initialize replay buffer R
4: **for** episode = 1, M **do**
5:    Initialize a random process N for action exploration
6:    Receive initial observation state $s_1$
7:    **for** t = 1, T **do**
8:       Select action $a_t = \mu(s_t|\theta^\mu)$ according to the current policy and exploration noise N
9:       Execute action $a_t$ and observe reward $r_t$ and observe new state $s_{t+1}$
10:      Store transition ($s_t$, $a_t$,$r_t$,$s_{t+1}$) in R
11:      Sample a random mini-batch of N transitions ($s_i$, $a_i$,$r_i$,$s_{i+1}$) from R
12:      Set $y_i = r_i + \gamma Q'(s_{i+1},\mu'(s_{i+1}|\theta^{\mu'})|\theta^{Q'})$
13:      Update critic by minimizing the loss:
$$L = \frac{1}{N}\sum_i (y_i - Q(s_{s_i}, a_i|\theta^Q))^2$$
14:      Update the actor policy using the sampled policy gradient:
$$\nabla_\theta J \approx \frac{1}{N} \sum_i \nabla_a Q(s, a|\theta^Q)|_{s=s_i, a=\mu(s_i)} \nabla_{\theta^\mu} \mu(s|\theta^\mu)|_{s_i}$$
15:      Update the target networks:
$$\theta^{Q'} \leftarrow \tau\theta^Q + (1 - \tau)\theta^{Q'}$$
$$\theta^{\mu'} \leftarrow \tau\theta^Q + (1 -\tau)\theta^{Q'}$$
16:
17:    **end for**
18: **end for**

The replay buffer is a finite sized buffer. At each time step, the actor and critic are updated by uniformly sampling a mini-batch from the replay buffer. Another addition in DDPG was the concept of *soft* target updates rather than directly copying the weights to the target network. The network Q(s,a|$\theta^Q$) being updated is also used to calculate the target value so the Q update is prone to divergence. This is possible by making a copy of actor and critic networks, Q'(s,a|$\theta^{Q'}$) and $\mu$'(s,a|$\theta^{\mu'}$). The weights of these networks are as : $\theta' \leftarrow \tau\theta + (1 - \tau)\theta'$ with $\tau \ll 1$. Another feature that DDPG derives from the technique in deep learning called batch normalization. This technique normalizes each dimension across the samples in a mini-batch to have unit mean and variance. The problem of exploration is solved by adding noise sampled from a noise process N to our actor policy.

$$\mu'(s_t) = \mu(s_t|\theta^\mu_t) + N \quad (5)$$

We have used an Ornstein-Uhlenbeck process with $\mu$ as 0.0, $\theta$ as 0.15 and $\sigma$ as 0.1 to generate temporally correlated exploration.

## III. MODELING AND SIMULATION

### A. Bipedal Walker

The proposed BWR was modelled in 3D modelling software, SolidWorks [see Fig.2]. The bipedal robot composed of five links - one waist section, two thighs and two shanks(Table.II). The dimensions of the links are depicted in the Fig.2. The bipedal walker was modelled so as to maintain centre of mass of the entire bipedal robot at the center the waist section. This maintains balance and is anthropomorphically similar to the humans. The specification of material chosen for each link is tabulated in Table.III. The range of rotation of the four joints i.e, hip and knee joints (both extension and flexion) during movement along sagittal plane is tabulated in Table.IV.

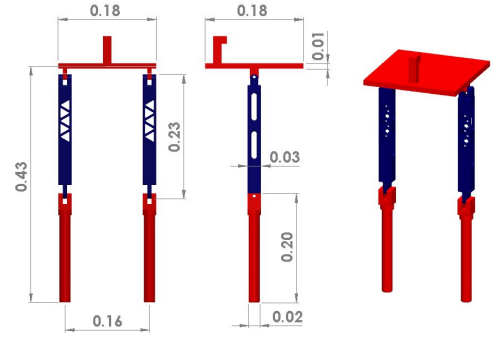

Fig. 3. CAD model of 5-Link Bipedal Walker in SolidWorks (All dimensions are in meters)

TABLE I
NOMENCLATURE OF THE LINKS OF BIPED

| Link Name | Terminology |
|---|---|
| Waist | waist |
| Right Thigh | thighR |
| Left Thigh | thighL |
| Right Shank | shankR |
| Left Shank | shankL |

TABLE II
LINK PROPERTIES OF THE BIPED WALKER.

| Link name | Material | mass(Kg) |
|---|---|---|
| waist | Acrylonitrile butadiene styrene(ABS) | 0.36416 |
| thighR | Acrylonitrile butadiene styrene(ABS) | 0.045155 |
| thighL | Acrylonitrile butadiene styrene(ABS) | 0.045155 |
| shankR | Acrylonitrile butadiene styrene(ABS) | 0.069508 |
| shankL | Acrylonitrile butadiene styrene(ABS) | 0.069508 |

TABLE III
ANTHROPOMETRIC DATA FOR THE MAXIMUM BENDING OF HIP AND KNEE JOINTS ALONG SAGITTAL PLANE.

| Joint | Flexion(radians) | Extension(radians) |
|---|---|---|
| Hip | 1.919862 - 2.26893 | 0.523599 |
| Knee | 2.26893 | 0.261799 |

### B. Simulation

The BWR was simulated in Gazebo, an open source 3D robotics simulator that is capable of recreating realistic environments for various robot based applications [23]. The 3D

TABLE IV
VARIOUS JOINT TYPES OF THE BIPED ROBOT.

| Parent Link | Child Link | Joint type |
|---|---|---|
| ground | stump | Fixed |
| stump | slider | Prismatic |
| slider | boom | Fixed |
| boom | waist | Prismatic |
| waist | thighR | Revolute |
| waist | thighL | Revolute |
| thighR | shankR | Revolute |
| thighL | shankL | Revolute |

CAD model of the biped walker, designed in SolidWorks was imported to the Gazebo simulator environment via a file conversion from .sldprt to .urdf formats. Unified Robot Description Format(URDF) is an eXtensible Markup Language (XML) file format is used to define the links and assemble them properly to recreate and render the robot in Gazebo environment. The URDF file of the robot model consists of each link's physical properties such as material, mass, length and moment of inertia. Also the location of origin(for each parent and corresponding child link) and axis of rotation for each link associated with the biped walker is defined in the URDF file. The joint types and positions for connecting the several links of the robot are specified through this format. The different joint types for connection of links are tabulated in Table.IV.

The connection of the links is in this order: Ground is connected to a cylindrical stump with a fixed joint. The stump is connected to a horizontal slider by a prismatic joint which in turn is connected to a boom. The waist's top is attached to a horizontal boom which slides forward and backwards along with the biped walker, to restrict the motion along sagittal plane(i.e, along Y-Z axis). The boom is of negligible mass when compared to the biped walker's and hence can be ignored. The visualization of the boom was neglected, to focus on the biped walker's interaction with the environment. Apart from these links, two contact sensors, one at the bottom of each shank was defined in the URDF file. This was to get the instants of contacts with the ground while walking. Hips rotation, Hips velocity, shank rotation, shank velocity, linear velocities in the sagittal plane and ground-foot contacts resulted into state space dimension of 12 and action space dimension of 4.

Robot Operating System (ROS) acted as an interface between the controller script and the Gazebo. The states were published on the respective topics and action commands were published back to control the links. The rate of communication between the script and Gazebo was 50 Hz.

## IV. RESULTS AND DISCUSSIONS

This section illustrates the results of the simulation of BWR to achieve stable walking gait. After training the bipedal walker on NVIDIA GeForce GTX 1050 Ti Graphical Processing Unit(GPU), for approximately 41 hours, a stable walking gait was achieved. The walker showed continuous forward walk for 10 meters without falling. The average reward per 100 episodes over the course of learning is shown in Fig. 4. The hip and knee rotation values are shown in Fig. 5 and Fig. 7 respectively. The results from the bipedal walking were compared with the actual human walking data captured using marker-based Optical Motion Capture System(mocap). To capture the motion, a motion capture suit was worn by a human subject. Markers were attached to the suit and the subject was asked to walk normally at his own pace. As can be seen from the Fig. 6 and Fig. 8, the characteristics of rotation angles recorded by mocap match with the rotation angles obtained during a bipedal walk. The hip rotations are approximately out of phase and the knee rotation frequency is double the hip rotation frequency.

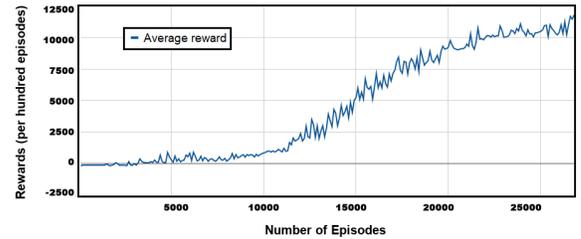

Fig. 4. Average reward per 100 episodes

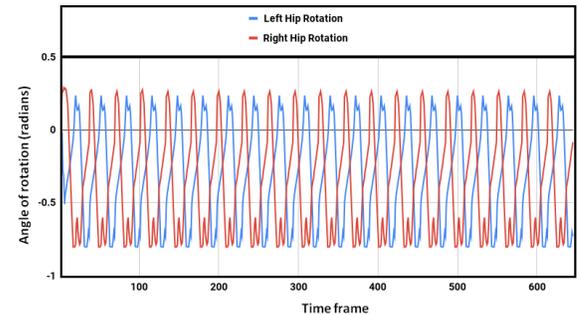

Fig. 5. Hip joints rotation trajectory of the BWR

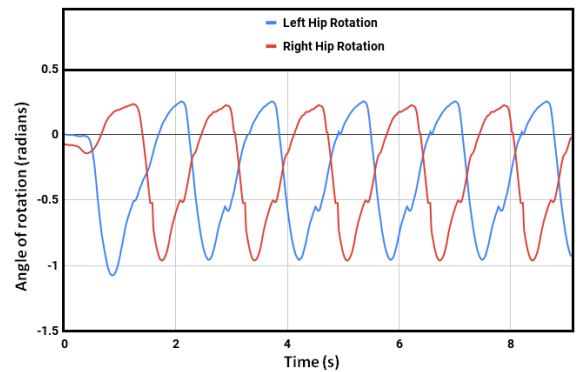

Fig. 6. Hip joints rotation trajectory of the human

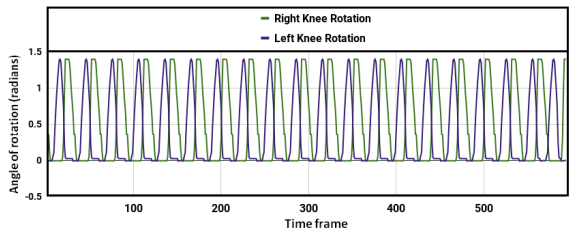

Fig. 7. Knee joints rotation trajectory of BWR

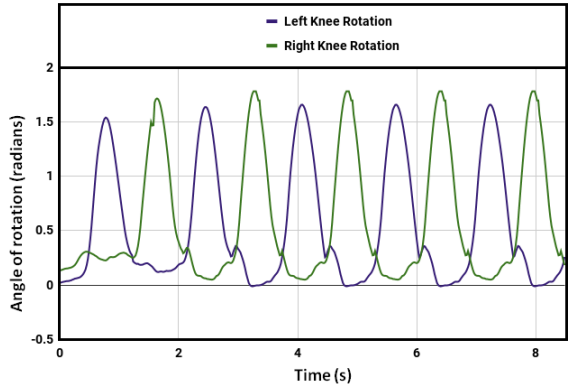

Fig. 8. Knee joints rotation trajectory of the human

## V. CONCLUSIONS

Reinforcement Learning can be used as a convenient method to learn complex controls without the prior knowledge of the dynamics of the agent or the environment. The same was demonstrated by the simulation of planar bipedal walker in real-world physics engine Gazebo. Over the recent years, more efficient reinforcement algorithms like Proximal Policy Optimization(PPO) and Trust Region Policy Optimization(TRPO) have been developed for optimal control. They have turned out to be better than the state of the art because of the ease of use and better performance. The future work will include these algorithms to better learn the bipedal walking. The efforts will also be focused towards hardware implementation and testing of the designed robot.Also, the current work tested the RL algorithms on the planar bipedal walker. In the future works, the two dimensional constraints will be eliminated and the algorithms will be tested in the three dimensional space.

## VI. ACKNOWLEDGEMENT

We would like to thank Dhanush GV for providing suggestions on initial BWR designs.We would also like to thank the Computational Intelligence Lab, Aerospace Engineering Department, IISc, Bangalore for providing the computation resources that were used to train the BWR model.